\documentclass[letterpaper,10pt,journal,twoside]{IEEEtran}

\usepackage{graphics} 
\usepackage{epsfig} 
\usepackage{times} 
\usepackage{amsmath} 
\usepackage{amssymb}  
\usepackage{hyperref}

\usepackage[textsize=footnotesize]{todonotes}

\usepackage[utf8]{inputenc} 
\usepackage{graphicx}
\usepackage{footnote}
\makesavenoteenv{tabular}
\makesavenoteenv{table}

\usepackage[ruled]{algorithm2e}

\usepackage{amsthm}
\usepackage{amsfonts}
\usepackage{mathtools}

\usepackage{enumitem}[nosep]
\usepackage{siunitx}

\usepackage{tabularx}
\usepackage{booktabs}

\usepackage{hyperref}  
\usepackage[capitalize,noabbrev]{cleveref}  
\crefformat{equation}{(#2#1#3)}
\hypersetup{%
    colorlinks=true,
    linkcolor=gray,
    urlcolor=gray,
    citecolor=gray,
    linktocpage=true
}
\usepackage[outdir=export/]{epstopdf}

\usepackage{multirow}
\usepackage{wrapfig}

\usepackage{algorithmic}
\usepackage{array}
\newcounter{rowcntr}[table]
\usepackage[font={small,it},belowskip=0pt]{caption}
\usepackage{placeins} 

\usepackage[normalem]{ulem}

\usepackage[percent]{overpic}
\usepackage{hyphenat}

\usepackage{xcolor}
\usepackage{bm}
\usepackage{textcomp}

\usepackage[nolist,nohyperlinks]{acronym}
\usepackage{mathtools}

\usepackage{adjustbox}
\usepackage{float}

\graphicspath{{drawings/}{export/}}

\newcommand{\ul}[1]{#1} 

\newcommand{\vc}[1]{\ensuremath{\bm{#1}}}

\newcommand{\bvec}[1]{\ensuremath{\begin{bmatrix}#1\end{bmatrix}}}
\usepackage{array}
\newcommand{\bvecs}[2]{\left[\hspace{-0.6em}\begin{tabular}{p{1.6cm}r}#1\end{tabular}\hspace{-0.6em}\right],}

\DeclarePairedDelimiter\abs{\lvert}{\rvert}
\DeclarePairedDelimiter\norm{\lVert}{\rVert}
\makeatletter
\let\oldabs\abs
\let\oldnorm\norm
\def\abs{\@ifstar{\oldabs}{\oldabs*}}
\def\norm{\@ifstar{\oldnorm}{\oldnorm*}}
\makeatother

\DeclareMathOperator{\rank}{rank}
\DeclareMathOperator{\trace}{tr}

\DeclareMathOperator{\Diag}{Diag}

\def\-{\raisebox{0pt}{-}}

\newcommand{\argmin}[1]{\underset{#1}{\operatorname{arg}\,\operatorname{min}}\;}

\newcommand{\R}[1]{\ensuremath{\boldsymbol{\mathbb{R}}^{#1}}}
\newcommand{\inR}[1]{\ensuremath{\in \R{#1}}}

\newcommand\kron{\ensuremath{\otimes}}

\theoremstyle{definition}

\let\oldunderbrace\underbrace
\renewcommand{\underbrace}[2]{\let\scriptstyle\textstyle \oldunderbrace{#1}_{#2}}
\definecolor{tab10blue}{rgb}{0.12156863, 0.46666667, 0.70588235}

\def\eg{\emph{e.g}.} 
\def\ie{\emph{i.e}.}

\DeclareFontFamily{U}{mathx}{\hyphenchar\font45}
\DeclareFontShape{U}{mathx}{m}{n}{ <-> mathx10 }{}
\DeclareSymbolFont{mathx}{U}{mathx}{m}{n}
\DeclareFontSubstitution{U}{mathx}{m}{n}
\DeclareMathAccent{\widebar}{\mathalpha}{mathx}{"73}
\newcolumntype{C}[1]{>{\centering\let\newline\\\arraybackslash\hspace{0pt}}m{#1}}
\newcolumntype{D}[1]{>{\centering\let\newline\\\arraybackslash\hspace{0pt}\columncolor{gray!20}}m{#1}}
\newcolumntype{R}[1]{>{\raggedleft\let\newline\\\arraybackslash\hspace{0pt}}m{#1}}
\newcolumntype{L}[1]{>{\raggedright\let\newline\\\arraybackslash\hspace{0pt}}m{#1}}
\newcolumntype{M}[1]{>{\raggedright\let\newline\\\arraybackslash\hspace{0pt}}m{#1}}
\newcolumntype{N}{>{\refstepcounter{rowcntr}\therowcntr}c}

\newcommand{\mytitle}[1]{\paragraph{#1}}

\newcommand{\new}[1]{#1}

\usepackage{tabstackengine}
\stackMath
\setstackgap{L}{18pt}
\setstacktabbedgap{0pt}
\def\lrgap{\kern6pt}

\newcommand{\hc}[1]{\ensuremath{\hat{\vc{#1}}}}

\newcommand\anchor{\ensuremath{\vc{y}}}
\newcommand\anchors{\ensuremath{\vc{Y}}}
\newcommand\Cov{\ensuremath{\vc{\Sigma}}}
\def\foralln{\ensuremath{\left(\forall n\right)}\quad}

  \renewenvironment{thebibliography}[1]{%
    \begin{oldthebibliography}{#1}%
      \setlength{\parskip}{0ex}%
      \setlength{\itemsep}{0ex}%
  }%
  {%
    \end{oldthebibliography}%
  }

\acrodef{QCQP}[QCQP]{quadratically-constrained quadratic program}
\acrodef{SDP}[SDP]{semidefinite program}
\acrodef{EVD}[EVD]{eigenvalue decomposition}
\acrodef{SVD}[SVD]{singular value decomposition}
\acrodef{MAP}[MAP]{maximum a posteriori}
\acrodef{GP}[GP]{Gaussian process}
\acrodef{LTV}[LTV]{linear, time-varying}
\acrodef{SLAM}[SLAM]{simultaneous localization and mapping}
\acrodef{IMU}[IMU]{inertial measurement unit}
\acrodef{psd}[PSD]{positive-semidefinite}
\acrodef{GN}[GN]{Gauss-Newton}
\acrodef{UWB}[UWB]{ultra-wideband}
\acrodef{rmse}[RMSE]{root-mean-squared error}
\acrodef{mae}[MAE]{mean absolute error}
\acrodef{LM}[LM]{Levenberg-Marquardt}
\acrodef{GPS}[GPS]{global positioning system}
\acrodef{AUV}[AUV]{autonomous underwater vehicle}

\title{Safe and Smooth: Certified Continuous-Time Range-Only Localization} 
\author{Frederike Dümbgen \quad Connor Holmes \quad Timothy D. Barfoot
    \thanks{This work was funded in part by the Swiss National Science Foundation, Postdoc Mobility under Grant 206954 and in part by the Natural Sciences and Engineering Research Council of Canada (NSERC).}
\thanks{The authors are with the University of Toronto Robotics Institute, University of Toronto, Toronto, M5S 1A4, Canada. Corresponding author: Frederike Dümbgen (frederike.dumbgen@utoronto.ca).}
}

\newcommand{\cor}[1]{#1}
\newcommand{\corn}[1]{#1}
\newcommand{\cornn}[1]{#1}
\newcommand{\corr}[1]{#1}
\renewcommand{\todo}[1]{}
\newcommand{\lt}[1]{#1}

\newcommand{\fracE}{\corr{\ensuremath{\frac{1}{E}}}}
\newcommand{\fracN}{\corr{\ensuremath{\frac{1}{N}}}}

\newcommand\myheader{\framebox{This paper was published in IEEE Robotics and Automation Letters, 2022 (\href{https://ieeexplore.ieee.org/document/10003973}{link}).}}
\usepackage{fancyhdr}
\fancypagestyle{firstpage}
{
  \fancyhead[L]{ \centering \myheader}
  \fancyhead[R]{}
}

\begin{document}

\maketitle 
\pagestyle{empty}

\renewcommand{\headrulewidth}{0pt}
\thispagestyle{firstpage}

\begin{abstract}
  A common approach to localize a mobile robot is by measuring distances to points of known positions, called anchors.  
  Locating a device from distance measurements is typically posed as a non-convex optimization problem, stemming from the nonlinearity of the measurement model. Non-convex optimization problems may yield suboptimal solutions when local iterative solvers such as Gauss-Newton are employed. 
In this paper, we design an optimality certificate for continuous-time range-only localization. 
Our formulation allows for the integration of a motion prior, which ensures smoothness of the solution and is crucial for localizing from only a few distance measurements. The proposed certificate comes at little additional cost since it has the same complexity as the sparse local solver itself: linear in the number of positions. We show, both in simulation and on real-world datasets, that the efficient local solver often finds the globally optimal solution (confirmed by our certificate), but it may converge to local solutions with high errors, which our certificate correctly detects.

\end{abstract}

\section{Introduction}

Localizing a moving robot is an essential component of many real-world applications. One common approach to localization\cor{, in particular when \ac{GPS} or cameras are unavailable,} is to measure the distances to a certain number of fixed points, called anchors \cor{or beacons}. In mobile indoor localization, for instance, a phone can be localized by inferring distances to WiFi access points from the time of flight or received signal strength of emitted pulses~\cite{raza_comparing_2021}. As another example, designated anchors equipped with the \ac{UWB} technology \corn{may be \cornn{used, for instance,} for autonomous lawnmowers \cornn{operating} in \cornn{environments} where feature-based computer vision methods are compromised~\cite{djugash_navigating_2009}, or for drones flying in \ac{GPS}-denied areas~\cite{shule_uwb-based_2020}.} \cor{Finally, in marine robotics, a common localization strategy \cornn{for autonomous submarines} is to emit sonar pulses and measure the time of flight to stationary beacons~\cite{olson_robust_2006}.}
In all these examples, the locations of the anchors are known a priori, or can be estimated in a separate procedure~\cite{hamer_self-calibrating_2018}. The remaining task is to determine a moving device's trajectory based on distance measurements, which \lt{is} also known as multilateration.

\begin{figure}[t]
  \centering
  \includegraphics[width=\linewidth]{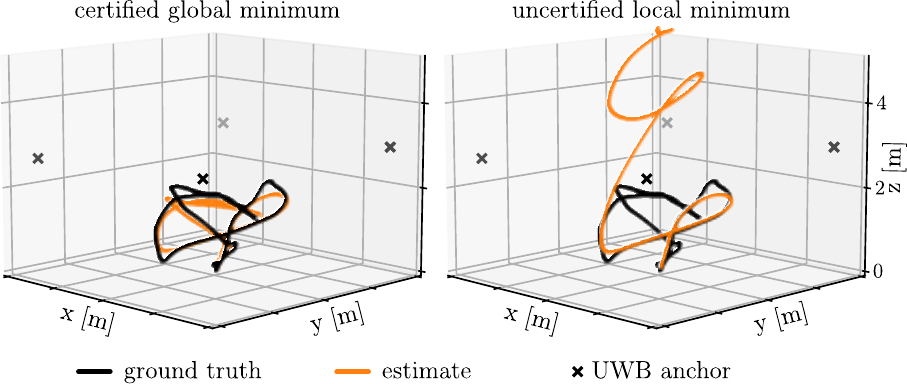}
  \caption{A flying drone measures distances to fixed anchor points in an arena. The two solutions are obtained by running a continuous-time range-only localization scheme from two different initializations. We propose an efficient optimality certificate based on Lagrangian duality that correctly identifies the left solution as the global minimum.}
  \label{fig:local-optimum-real}
\end{figure}

Although multilateration \corn{has been studied for a long time}, many important open questions persist. For instance, when approached from a robotics point of view, multilateration is called range-only localization and typically involves solving a nonlinear least-squares optimization problem \cornn{with, at best, local convergence guarantees}~\cite{nocedal_numerical_2006}. A more \cornn{optimal} approach to multilateration is to exploit principles from distance geometry~\cite{liberti_euclidean_2014}. \cornn{However, optimality and recovery guarantees obtained this way usually assume no noise~\cite{so_theory_2007}, or that each position can be uniquely localized~\cite{larsson_optimal_2019,beck_exact_2008}.}
Figure~\ref{fig:local-optimum-real} displays the limitation of these existing methods. \cor{In the shown example, a drone is localized based on distance measurements to fixed~\ac{UWB} anchors. The distance measurements are noisy \cornn{and sparse (we only measure one distance at a time)}, which rules out optimal solvers from distance geometry~\cite{larsson_optimal_2019,beck_exact_2008}. Instead, we can employ a continuous-time range-only localization framework \cornn{with a local solver~\cite{barfoot_state-estimation}.} However, as the example shows, such a method may yield a suboptimal solution far from the global optimum, if poorly initialized.}

The method presented in this paper allows us to \cor{efficiently identify optimal solutions}. We first derive a certificate for discrete-time range-only localization without between-point dependencies, \cornn{which can be used when we measure enough distances at each time. By incorporating smoothness priors, we extend the operating conditions of both solver and certificate to cases where only a few distance measurements are available at each time.}
\cor{This is allowed by \corr{posing} the problem as continuous-time range-only localization, for which no optimal solvers are known to this date. In the continuous-time framework,} smoothness priors are enforced by regularization terms that stem from~\ac{GP} regression and allow us to incorporate physical assumptions about the trajectory. As a welcome side effect, the continuous model can be used to \cornn{interpolate} the trajectory, or to obtain \cor{closed-form} estimates of quantities of interest such as the instantaneous velocity~\cite{barfoot_batch_2014}. The proposed certificates show that \new{local solvers} yield the optimal solution in the vast majority of cases. More specifically, the contributions of this paper are:
\begin{itemize}
  \item certificates for multilateration and range-only continuous-time localization,

  \item a detailed treatment of how to exploit sparsity to \new{compute the certificates in linear complexity}, and
  \item validation of the proposed certificates, including evaluation on real-world datasets.
\end{itemize}

This paper is organized as follows. After reviewing related work in Section~\ref{sec:related-work}, we introduce the certificate for multilateration and continuous-time range-only localization in Sections~\ref{sec:methods-multi} and~\ref{sec:methods-ro}, respectively. We include a discussion of our local solver in Section~\ref{sec:solver} and the certificate computation in Section~\ref{sec:sparsity}.
We evaluate the certificates in simulation and on a \ac{UWB}-based drone localization \cor{dataset} in Section~\ref{sec:results}, and conclude in Section~\ref{sec:conclusion}.

\section{Related Work}\label{sec:related-work}

\textbf{Range-only localization} 
The underdetermined nature of distance measurements makes range-only localization \cor{a} challenging subclass of state estimation problems. \cor{When the anchors are unknown and to be estimated along with the moving device, we \corn{refer to this as} range-only \ac{SLAM}.} To account for the multi-modal distribution stemming from underdetermined measurements, prior work has focused on using Gaussian mixture models~\cite{fabresse_undelayed_2013, vallicrosa_sum_2016} or sample-based models~\cite{blanco_efficient_2008,blanco_pure_2008} to approximate the position distribution in a filtering-inspired framework. More accurate than filters are batch solutions, which typically provide the \ac{MAP} estimate of the trajectory given all measurements within a given time window~\cite{barfoot_state-estimation}. \cornn{In general}, batch solutions are more expensive than filtering, but the sparsity of the underlying measurement graph typically allows for efficient and incremental sparse solvers~\cite{dellaert_square_2006,Kaess2012}. In both batch and filter approaches, continuous-time rather than discrete-time trajectory models have been explored~\cite{barfoot_state-estimation}, through which smoothness can be incorporated, thereby helping with underdetermined measurements. To make the continuous-time estimation tractable, some prior work uses parametric representations with carefully chosen temporal basis functions~\cite{pacholska_relax_2020,furgale_continuous-time_2012}. A popular alternative is non-parametric \ac{GP} regression, which is easier to tune and has an elegant connection with physically plausible motion priors~\cite{barfoot_batch_2014}.

Whatever representation is used, at the core of batch \ac{MAP} estimation is the solution of a non-convex minimization problem~\cite{barfoot_state-estimation}. The latter is most commonly solved with an iterative minimizer, which \corn{usually} converges to a stationary point, but not necessarily to the global minimum~\cite{nocedal_numerical_2006}.
Different approaches originated in the sensor network localization literature and study semi-definite relaxations and optimality guarantees~\cite{biswas_semidefinite_2004-1,biswas_semidefinite_2006}. However, optimality guarantees only exist for the noiseless case~\cite{so_theory_2007}, and this approach scales poorly with batch size. Similarly, works inspired by distance geometry \cornn{(see~\cite{liberti_euclidean_2014,dokmanic_euclidean_2015} for overviews of the topic)} can be exploited but, again, efficient algorithms with recovery guarantees \cornn{do not allow for noisy~\cite{so_theory_2007} or under-determined problems~\cite{larsson_optimal_2019,beck_exact_2008}}.

With the proposed solution, we can be both efficient and optimal: we use a continuous-time batch approach, but exploit sparsity to keep the cost low enough for online inference, both in solving and certifying the solution. We place no assumptions on noise or uniqueness of the solution -- as long as the certificate holds, the solution is optimal.

\textbf{Optimality certificates}
In the last 10 years, significant progress was made in certifying solutions of common optimization problems in robotics and related fields. Certificates typically originate from Lagrangian duality principles~\cite{boyd_convex_2004}, and were primarily introduced to robotics and computer vision for the problems of pose-graph optimization~\cite{carlone_lagrangian_2015,carlone_duality-based_2015} and rotation averaging~\cite{fredriksson_simultaneous_2013}, respectively. The main difficulty of these problems stems from the non-convex constraints that emerge when estimating rotations. Existing provably optimal solvers require the solution of a~\ac{SDP}, which extends poorly to large batch sizes. Instead, follow-up works have investigated careful reformulation~\cite{briales_fast_2016}, Riemannian optimization~\cite{rosen_se-sync_2019}, block coordinate descent~\cite{eriksson_rotation_2018}, and an adaptation of, for instance, \ac{GN}~\cite{dellaert_shonan_2020}, to speed up the certified solvers. In parallel strands of research, significant progress has been made on using outlier-robust cost functions for optimal line fitting~\cite{fineline} and optimal point-cloud alignment~\cite{yang_graduated_2020-1,yang_teaser_2020}, respectively.
\corn{Very recently, the above ideas have been applied to the joint localization of static points from their inter-point distances~\cite{halsted_riemannian_2022,maric_riemannian_2022}. Instead, we consider the widespread anchor-based localization. Furthermore, our method allows for the use of motion priors, crucial for localization in real-world settings where distance measurements may be sparse and corrupted by high noise.}

\section{Method}\label{sec:methods}

\subsection{\lt{Preliminaries}}

Our goal is to solve for an unknown state vector over time, which we denote by $\vc{\theta}(t)\inR{K}$ at time $t$. The state typically contains the \cornn{robots'} position $\vc{x}(t)\inR{D}$, with $D$ equal to 2 or 3\cornn{.} As we see later, extending the state to include, for instance, the velocity, gives us more flexibility for imposing \cornn{motion} priors. At \cornn{given} times $t_n, n=1,\ldots,N$, we obtain distance measurements $d_{mn}$ from the position $\vc{x}_n:=\vc{x}(t_n)$ to known anchors $\anchor_m \inR{D}, m=1,\ldots,M$. \cornn{The number of observed anchors from position $n$ is denoted by $M_n$, with $E=\sum_n M_n$ the total number of measurements. The measurement model is thus}
\begin{equation}
  \cor{\vc{h}_n(\lt{\vc{\theta}_n}) =}
  \bvec{\norm{\anchor_1 - \vc{x}_n}^2 \\
    \vdots \\
    \norm{\anchor_{M_n} - \vc{x}_n}^2 
  }\cor{=\bvec{d_{1n}^2 \\ \vdots \\ d_{M_n n}^2},}
  \label{eq:measurement-model}
\end{equation}
\cor{\lt{with $\vc{\theta}_n:=\vc{\theta}(t_n)$ and $d_{mn}$ the distance from anchor $m$ to position $n$.} We denote the noisy measurements by $\widetilde{\vc{d}}_n:=\vc{h}_n\left({\vc{\theta}_n}\right)+\vc{n}_n$}, with $\vc{n}_n$ \corn{measurement noise with covariance $\vc{\Sigma}_n$}. We introduce the matrix of known anchor coordinates $\anchors_n=\bvec{\anchor_1\cdots\anchor_{M_n}}\inR{D \times M_n}$ and the vector of its squared norms $\vc{\gamma}^\top_n = \bvec{\norm{\anchor_1}^2 \cdots \norm{\anchor_{M_n}}^2}$ \inR{M_n}. $\vc{I}_d$ and $\vc{1}_d$ are the $d$-dimensional identity matrix and the vector of all ones, respectively, and $\vc{i}_n^d$ is the length-$d$ selection vector with a one at index $n$. Finally, we write all time-concatenated vectors as $\vc{x}^\top=\bvec{\vc{x}_1^\top \cdots \vc{x}_N^\top}$, and the block-diagonal matrix composed of elements $\vc{A}_n$ as $\Diag(\vc{A}_n)_{n=1}^N$. The Kronecker product is written as $\kron$\lt{, and $\vc{A}\succeq 0$ signifies $\vc{A}$ is a \ac{psd} matrix.}  

\subsection{Certified Multilateration}\label{sec:methods-multi}

We first derive a certificate for range-only localization without imposing any smoothness on the trajectory. Therefore, we assume that our state is discrete and consists only of the position: $\lt{\vc{\theta}_n}=\vc{x}_n\inR{D}$. The problem could be separated into $N$ smaller problems, each of which has an optimal solution~\cite{larsson_optimal_2019,beck_exact_2008}, but treating it jointly serves as a convenient starting point for the continuous-time certificates.

\paragraph{Problem Statement}

The \ac{MAP} estimate can be obtained by solving 
\begin{align}\label{eq:cost-ro}
  \hc{\theta} &= \argmin{\vc{\theta}} f(\vc{\theta}) = \argmin{\vc{\theta}} \fracE\sum_{n} \vc{e}_n^\top \Cov_n^{-1} \vc{e}_n,
\end{align}
with $\vc{e}_n = \cor{\widetilde{\vc{d}}_n} - \vc{h}_n(\vc{\theta}_n)$.\footnote{\cor{We use $\vc{\theta}$ instead of $\vc{x}$, although the state only contains the position, to make the extension to the next Section more apparent.}} Expanding row $m$ of the error vector $\vc{e}_n$ yields
\begin{equation}
  \begin{aligned}
    e_{mn} &= \cor{\widetilde{d}_{mn}^2} - \norm{\anchor_m - \vc{x}_n}^2 \\
    &= \cor{\widetilde{d}_{mn}^2} - \norm{\anchor_m}^2 + 2\anchor_m^\top\vc{x}_n - \norm{\vc{x}_n}^2 , 
  \end{aligned}
  \label{eq:error}
\end{equation}
which shows that~\eqref{eq:cost-ro} is a quartic function in the unknown vectors $\vc{x}_n$. To turn it into a quadratic function, \lt{we introduce the substitution} $z_n=\norm{\vc{x}_n}^2$. Then, \lt{using} $\vc{f}_n^\top:=\bvec{\vc{x}_n^\top & z_n}$ and $\vc{b}_{n} := \cor{\widetilde{\vc{d}}_{n}} - \vc{\gamma}_n$ we can rewrite the error vector as $\vc{e}_n = \vc{Q}_n \vc{f}_n + \vc{b}_n$, with $\vc{Q}_n := \bvec{2\anchors_n^\top & -\vc{1}}$. 
We obtain the following \ac{QCQP}, equivalent to~\eqref{eq:cost-ro}:
\begin{align}
  \min_{\vc{f}_n, n=1,\ldots,N} &\fracE\sum_n (\vc{Q}_n^\top \vc{f}_n + \vc{b}_n)^\top \Cov_n^{-1} (\vc{Q}_n^\top \vc{f}_n + \vc{b}_n)  \label{eq:qcqp-1} \\
  \text{s.t. } \vc{f}_n^\top &\begin{bmatrix} \vc{I}_d & \vc{0} \\ \vc{0}^\top & 0 \end{bmatrix} \vc{f}_n - \begin{bmatrix} \vc{0}^\top & 1 \end{bmatrix} \vc{f}_n = 0 \quad n=1,\ldots,N. \nonumber
\end{align}
Introducing $\vc{f}\inR{F}$, the vector of stacked variables of size $F=N(D+1)+1$,
\begin{equation}\label{eq:traj-f}
  \vc{f}^\top=\bvec{\vc{x}_1^\top & z_1 & \corn{\cdots} & \vc{x}_N^\top & z_N & \ell},
\end{equation}
with $\ell$ a homogenization variable, we can convert \eqref{eq:qcqp-1} into the standard, homogeneous \ac{QCQP}: 
\begin{equation}\label{eq:traj-qcqp}
\begin{aligned}
  \text{(Q)} \quad q^* = &\min_{\vc{f}} \fracE \vc{f}^\top \vc{Q} \vc{f} \\
  \text{s.t. } &\vc{f}^\top \vc{A}_n \vc{f} = 0 \quad n=1,\ldots,N\\
  &\vc{f}^\top {\vc{A}_0 \vc{f}} = 1,\\
\end{aligned}
\end{equation}
with $\vc{A}_0\inR{F\times F}$ the all-zero matrix except for a one for the bottom-right element, and  $\vc{A}_n,\vc{Q} \inR{F\times F}$ given by
\vspace{-0.4em}
\begin{equation}
  \vc{A}_n=\vc{S}_n \begin{bmatrix} \vc{I}_d & \vc{0} & \vc{0} \\ \vc{0}^\top & 0 & -\frac{1}{2} \\ \vc{0}^\top & -\frac{1}{2} & 0 \end{bmatrix} \vc{S}_n^\top,\; \vc{S}_n=\bvec{\vc{i}_n^N \kron \vc{I}_{d+1} & \vc{0} \\ \vc{0}^\top & 1},
\end{equation}

\begin{equation*}
  \begin{aligned}
    \vc{Q} &= \bvec{
	\vc{Q}_{11} & \cdots & \vc{0} & \vc{q}_1 \\
	\vdots & \ddots & \vdots & \vdots \\
  \vc{0} & \cdots & \vc{Q}_{NN} & \vc{q}_N \\
\vc{q}_1^\top & \cdots & \vc{q}_N^\top & q_0}, \; q_0 = \sum_n \vc{b}_n^\top\Cov_n^{-1}\vc{b}_n,\\
\vc{Q}_{nn} &= \bvecs{
      $4\anchors_n\Cov_n^{-1}\anchors_n^\top$ & $-2\anchors_n\Cov_n^{-1}\vc{1}$  \\
    $-2\vc{1}^\top\Cov_n^{-1} \anchors_n^\top$ &$ \vc{1}^\top\Cov_n^{-1}\vc{1}$} ,\;
    \vc{q}_n = \bvec{2\anchors_n\Cov_n^{-1}\vc{b}_n \\ -\vc{1}^\top \Cov_n^{-1}\vc{b}_n}.
  \end{aligned}
  \label{eq:Q}
  \end{equation*}
\noindent
\cor{\ac{QCQP} problems are non-convex and may be hard to solve optimally~\cite{boyd_convex_2004}. However,~\corn{it is well known that we can exploit semidefinite relaxations, followed by Lagrangian duality theory}, to analyze (or potentially compute globally optimal) solutions (see, \eg,~\cite{eriksson_rotation_2018,rosen_se-sync_2019,cifuentes_local_2022}). Using this paradigm}, we turn problem (Q) into a \ac{SDP} by introducing $\vc{F}=\vc{f}\vc{f}^\top$ (which is equivalent to $\vc{F}\succeq0$, $\rank{\vc{F}}=1$) and relaxing the rank constraint. This gives the standard \ac{SDP} relaxation of (Q), which we denote as the primal problem (P)~\cite{boyd_convex_2004}:
\begin{equation}
  \begin{aligned}\label{eq:traj-primal}
    \text{(P)} \quad p^* = &\min_{\vc{F}} \fracE \trace{\left(\vc{Q}^\top \vc{F}\right)} \\
    \text{s.t. } &\trace{\left(\vc{A}_n \vc{F}\right)} = 0 \quad n=1,\ldots,N\\
    &\trace{\left(\vc{A}_0 \vc{F}\right)} = 1\\
    &\vc{F}\succeq 0.
\end{aligned}
\end{equation}
The \cor{(Lagrangian)} dual problem~\cite{boyd_convex_2004} is given by
\begin{equation}
\begin{aligned}\label{eq:traj-dual}
  \text{(D)} \quad d^* = &\max_{\rho, \vc{\lambda}}( -\rho) \\
  \text{s.t. }&  \vc{H}(\rho, \vc{\lambda}) := \fracE \vc{Q} + \sum_n \lambda_n \vc{A}_n + \rho \vc{A}_0 \succeq 0,
\end{aligned}
\end{equation}
where $\rho$ and $\vc{\lambda}^\top=\bvec{\lambda_1\cdots\lambda_N}\inR{N}$ are \cor{called the Lagrange multipliers or} dual variables.

At this point, it is useful to take a step back and consider what we have achieved so far. We have relaxed our original nonlinear state estimation into a standard \ac{SDP}. Now, we could take at least two different approaches to solving the original problem:
\begin{itemize}
  \item Solve the relaxed problem (P), and investigate the solution, denoted by $\vc{F}^*$. If the obtained solution has rank $1$, then we can decompose it into $\vc{F}^*=\vc{f}^*{\vc{f}^*}^\top$, where $\vc{f}^*$ is exactly the globally optimal solution to (Q). 
  \item Solve the primal problem (Q) locally using an iterative nonlinear solver. This will return a solution that is ensured to be locally optimal; we call this estimate $\hc{f}$. Then, we can use optimality conditions from duality theory to derive a certificate for this solution: if the certificate holds, the solution is in fact optimal and we have $\hc{f}=\vc{f}^*$.
\end{itemize}
In this paper, we take the second approach. This choice is motivated by two observations. First, in standard localization problems, we aim to solve for a large number of points simultaneously, yielding a large \ac{SDP} for (P). The typical complexity of available solvers is cubic in the number of points~\cite{nocedal_numerical_2006}, making them too slow for real-time robotics applications. Second, we found that even a basic iterative solver often converges to the optimal solution, in particular for the noise levels that are adequate for localization problems. As we will show, such solvers can exploit the sparsity of the problem in a principled manner, which makes them significantly faster than \ac{SDP} solvers. \corn{A similar approach was used in recent works to certify the solutions of other common problems in robotics such as pose-graph optimization~\cite{rosen_se-sync_2019} and landmark-based \ac{SLAM}~\cite{rosen_se-sync_2019,holmes_efficient_2022}.  \cornn{These problems assume linear measurement models, while we treat non-linear distance measurements.}}

\paragraph{Certificate}

Our aim is to determine whether a locally optimal solution $\hc{f}$ \cornn{to (Q)} is also the global optimum. We obtain a local solution $\hc{x}$ \cornn{to~\eqref{eq:cost-ro}} from a standard iterative~\ac{GN} solver, as outlined in~\ref{sec:solver}, and augment it to \hc{f} as in~\eqref{eq:traj-f}. We know from duality theory (see~\eg,~\cite{cifuentes_local_2022}) that if we can find dual variables $\hat{\rho},\hc{\lambda}$ such that:
\begin{align}
  &\hc{f}^\top\vc{A}_0\hc{f}=1, (\forall n)\; \hc{f}^\top\vc{A}_n \hc{f}=0  \;\; \text{(primal feasibility),} \label{eq:primal-feas}\\
&\vc{H}(\hat{\rho},\hc{\lambda}) \succeq 0 \quad \text{(dual feasibility), and} \label{eq:dual-feas}\\
&\vc{H}(\hat{\rho},\hc{\lambda})\hc{f}=\vc{0} \quad \text{(stationarity condition),} \label{eq:stationarity}
\end{align}
then $\hc{f}$ (and thus $\hc{x}$) is in fact the optimal solution to (Q). 
Because $\vc{H}(\hat{\rho}, \hc{\lambda})$ plays such a crucial role, we will refer to it as the `certificate matrix'.
Note that the conditions are sufficient, but not necessary -- if a solution does not satisfy all conditions it may still be an optimal solution. \cor{However, related works have shown that for sufficiently low noise levels, \corn{strong duality holds}, and the certificate becomes both sufficient \textit{and} necessary~\cite{cifuentes_local_2022}. This is confirmed and further discussed in the simulated experiments in Section~\ref{sec:simulation}}. 

The primal feasibility is trivial by construction, so we only need to verify the last two conditions. We can rewrite the stationarity condition as
\begin{equation}\label{eq:lin-sys-here}
  \bvec{\vc{A}_1\hc{f} & \cdots & \vc{A}_N\hc{f} & \vc{A}_0\hc{f}} \hc{y} = -\fracE\vc{Q}\hc{f}\text{,}
\end{equation}
with $\hc{y}:=\bvec{\hat{\lambda}_1 & \cdots & \hat{\lambda}_N & \hat{\rho}}^\top$, which is a linear system with $F$ equations and $N+1$ unknowns, and a priori, may not have a solution. However, \cor{we show in \corn{the supplementary material}~\cite{safe_arxiv} that the system admits the \cor{unique} solution:}
\cor{
\begin{align}
  \foralln \hat{\lambda}_n &= -2\fracE\vc{1}^\top \Cov_n^{-1}\vc{e}_n, \label{eq:lambda} \\
  \hat{\rho} &= -\fracE\sum_n \vc{e}_n^\top\Cov_n^{-1}\vc{e}_n. \label{eq:rho}
\end{align}
\noindent Note }that the analytical solution of $\hat{\rho}$ shows that strong duality holds between (Q) and (D) \cornn{provided~\eqref{eq:dual-feas} is true, as we have $d^* = -\hat{\rho} = \cor{q^*}$.}

In summary, given a locally optimal solution $\hc{f}$, we can use~\eqref{eq:lambda} and~\eqref{eq:rho} to solve for the optimal dual variables. If they are such that the certificate matrix is \ac{psd}, all conditions~\eqref{eq:primal-feas} to~\eqref{eq:stationarity} are satisfied and we conclude that $\hc{f}$ is in fact the optimal solution to (P). 

\subsection{Certified Continuous-Time Range-Only Localization}\label{sec:methods-ro}

The previous example has taught us how to certify the optimality of a candidate solution to the range-only localization problem. However, we reiterate that efficient optimal solvers for problem~\eqref{eq:cost-ro} exist, therefore locally solving and then certifying the solution is not \cornn{necessary} in general. 

However, we can use what we have learned to extend the method to incorporate motion priors, a case for which no efficient, provably optimal solvers are known. \cor{Instead of an optimal solver,} the certificates developed hereafter allow us to use a local solver, followed by a certificate check, both of which can be implemented efficiently by exploiting sparsity. As we will see in Section~\ref{sec:results}, the fast local solver finds the global optimum most of the time, and suboptimal solutions can be avoided through simple reinitialization; our certificate can tell us when this is necessary.

\paragraph{Motion Prior}

Since robots move according to physical laws, the resulting trajectories typically exhibit a certain degree of smoothness. A principled way to formalize this fact is by expressing the trajectory as a \ac{GP}, with a one-to-one correspondence between both the covariance and mean functions and the \cornn{motion prior. Following the method outlined in~\cite{barfoot_batch_2014}, we show in this Section that we can incorporate a motion prior in our \ac{MAP} estimation by solving:
\begin{equation}\label{eq:cost-gp}
  \begin{aligned}
  \hc{\theta} &= \argmin{\vc{\theta}} f(\vc{\theta}) + r(\vc{\theta}),
  \end{aligned}
\end{equation}
where $f(\vc{\theta})$ is defined in~\eqref{eq:cost-ro}, and $r(\vc{\theta})$ is a regularization term enforcing the prior. In the following sections, we will show that we can use a similar methodology as in multilateration to certify solutions to problem~\eqref{eq:cost-gp}. 
}

In contrast with Section~\ref{sec:methods-multi}, the state vector $\vc{\theta}(t) \inR{K}$ may now consist of more states than just the position; for instance, we may add \cor{velocity and even acceleration}. We assume that $\vc{\theta}(t)$ is a \ac{GP}:  
\begin{equation}
  \vc{\theta}(t) \sim \mathcal{GP}\left(\vc{\mu}(t), \vc{K}(t, t')\right), \quad t_0 < t, t'
  \label{eq:gp}
\end{equation}
where $\vc{\mu}(t)$ is the mean function, $\vc{K}(t, t')$ is the covariance function between two times, and $t_0$ is the starting time. \corn{We set the prior mean function to zero, as is commonly done in practice.} One particular class of covariance functions comes from \cornn{assuming a} \ac{LTV} system: 
\begin{equation}
  \dot{\vc{\theta}}(t) = \vc{A}(t)\vc{\theta}(t) + \vc{B}(t)\vc{u}(t) + \vc{F}(t)\vc{w}(t),
\end{equation}
\new{with $\vc{A}(t)$, $\vc{B}(t)$ and $\vc{F}(t)$ known system matrices,} $\vc{u}(t)$ a known input and $\vc{w}(t)\sim\mathcal{GP}\left(\vc{0}, \vc{Q}_C\delta(t - t')\right)$, a stationary zero-mean \ac{GP} with power spectral density matrix $\vc{Q}_C$.
The general solution to this model is
\begin{equation}
  \begin{aligned}
  \vc{\theta}(t) &= \vc{\Phi}(t, t_0)\vc{\theta}(t_0) + \\
  &\int_{t_0}^t\vc{\Phi}(t, s)\left(\vc{B}(s)\vc{u}(s) + \vc{F}(s)\vc{w}(s)\right)ds,
  \end{aligned}
\end{equation}
where $\vc{\Phi}(t,t')$ is the transition function. To make the model more tangible, we introduce two example motion priors. 

\textbf{Example 1: zero-velocity prior} By setting the state to the position only ($\vc{\theta}(t) =\vc{x}(t)$), and the system matrices to $\vc{A}(t)=\vc{F}(t)=\vc{I}$, and $\vc{B}(t)=\vc{0}$, we obtain the `zero-velocity' prior, meaning we assume that there is no motion between two consecutive points. In this case, the transition matrix is simply $\vc{\Phi}(t, t')=\vc{I}$, and we obtain a regularization term equivalent to Tikhonov regularization~\cite{tikhonov_solutions_1979}. 

\textbf{Example 2: constant-velocity prior} \cornn{The} constant-velocity assumption \cornn{is imposed} by setting:
\begin{align*}
  \vc{\theta}(t)=\bvec{\vc{x}(t)\\ \vc{v}(t)}, \vc{A}(t)=\bvec{\vc{0} & \vc{I} \\ \vc{0} & \vc{0}}, \vc{B}(t)=\vc{0}, \vc{F}(t)=\bvec{\vc{0} \\ \vc{I}},
\end{align*}
where $\vc{v}(t)$ denotes the velocity at time $t$. In this case, the transition matrix is
\begin{equation}
  \vc{\Phi}(t, t') = \bvec{\vc{I} & (t - t')\vc{I} \\ \vc{0} & \vc{I}}.
  \label{eq:phi}
\end{equation}

\begin{figure}[tb]
  \centering
  \includegraphics[width=\linewidth]{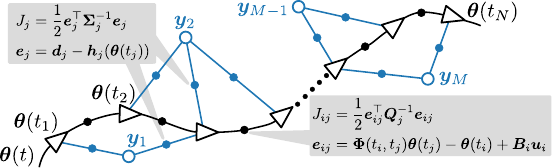}
  \caption{Factor graph representation of the \ac{GP} inference problem. Black factors represent the motion prior, blue are the range-only measurement factors.}
  \label{fig:factor-graph}
\end{figure}

\noindent
\corn{Any higher-order priors from the \corn{\ac{LTV} family}, such as white-noise-on-jerk (constant-acceleration) \lt{priors}, can be derived analogously~\cite{barfoot_batch_2014}. \corn{For all of these priors,} the regularization term takes the form}
\begin{equation}
  \quad r(\vc{\theta}) := \fracN \sum_{n=2}^{N} \vc{e}_{n,n-1}\vc{Q}_{n}^{-1} \vc{e}_{n,n-1},
\end{equation}
where we have introduced
\begin{align}
    \vc{e}_{n,n-1} &:= \vc{\Phi}_{n,n-1}\vc{\theta}_{n-1} - \vc{\theta}_{n} + \vc{B}_n \vc{u}_{n} \\
    \vc{u}_n:&=\int_{t_{n-1}}^{t_n} \vc{\Phi}(t_n,s)\vc{B}(s)\vc{u}(s)ds \\ 
    \vc{Q}_n:&=\int_{t_{n-1}}^{t_n} \vc{\Phi}(t_n,s)\vc{F}(s)\vc{Q}_c{\vc{F}(s)}^\top{\vc{\Phi}(t_n,s)}^\top ds , 
  \end{align}
  and $\vc{\theta}_n:=\vc{\theta}(t_n)$, \new{$\vc{B}_n:=\vc{B}(t_n)$, and} $\vc{\Phi}_{ij}:=\vc{\Phi}(t_i, t_j)$. The important point to note is that each regularization term in~\eqref{eq:cost-gp} depends only on \cor{two adjacent states}, owing to the Markov property. The inference problem is thus still sparse, which we will exploit later to develop efficient solvers. To visualize this point, we show in Figure~\ref{fig:factor-graph} the factor-graph representation of the inference problem. The \cor{\ac{GP} prior} results in state-to-state factors, taking a similar role as odometry measurements would in classical \ac{SLAM} problems.  

\paragraph{\cor{Problem Statement}}

\begin{figure}
    \centering
    \includegraphics[width=\linewidth]{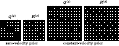}
    \caption{Sparsity patterns of the cost matrices obtained using zero-velocity or constant\cor{-}velocity priors, for \cor{$N=3$} and $D=3$.}
    \label{fig:cost-matrices}
\end{figure}

\cor{We have now derived all ingredients to bring the estimation problem with motion prior into a standard \ac{QCQP}.} First, we note that $r(\vc{\theta})$ can be written as \corn{
\begin{equation}
  r(\vc{\theta}) = \fracN \vc{\theta}^\top \vc{R} \vc{\theta}, 
\end{equation}
where $\vc{R}$ is the block-tridiagonal matrix with off-diagonal elements}
$\vc{R}_{n,n+1} = - \vc{\Phi}_{n+1,n}^\top\vc{Q}_{n+1}^{-1}$  for $1 \leq n \leq N\- 1$ and diagonal elements \\ 
\begin{equation}
  \begin{aligned}
  &\vc{R}_{nn} = \begin{cases}
    \vc{Q}_{n}^{-1} + \vc{\Phi}^\top_{n+1,n}\vc{Q}_{n+1}^{-1}\vc{\Phi}_{n+1,n} &\text{for } 2 \leq n \leq N\- 1 \\
     \vc{\Phi}_{n+1,n}^\top\vc{Q}_{n+1}^{-1}\vc{\Phi}_{n+1,n} &\text{for } n=1 \\ 
     \vc{Q}_n^{-1} &\text{for } n=N
    \end{cases}.
  \end{aligned}
  \label{eq:R}
\end{equation}
\noindent
We also introduce the new vector $\vc{g}^\top = \bvec{\vc{\theta}_1^\top & z_1 &\cdots & \vc{\theta}_N^\top & z_N & 1}$
where we do not add any substitution variables for the components of $\vc{\theta}_n$ other than $\vc{x}_n$ because the added regularization is already quadratic in $\vc{\theta}_n$.  We then augment $\vc{R}$ with zero rows and columns where the regularization is zero (\ie, for the substitution variables), yielding $\vc{R}^{(g)}$. Similarly, we create $\vc{Q}^{(g)}$ and $\vc{A}_i^{(g)}$, $i=0,\ldots,N$ by padding with zeros for the variables in $\vc{\theta}_n$ other than the position. Finally, we can write~\eqref{eq:cost-gp} in the standard form
\begin{equation}\label{eq:gp-qcqp}
\begin{aligned}
  \text{(Q-GP)} \quad q_{GP}^* = &\min_{\vc{g}} \fracE \vc{g}^\top \vc{Q}^{(g)} \vc{g} + \fracN \vc{g}^\top \vc{R}^{(g)} \vc{g} \\
  \text{s.t. } &\vc{g}^\top \vc{A}_{n}^{(g)} \vc{g} = 0 \quad n=1,\ldots,N \quad \\
  &\vc{g}^\top {\vc{A}_{0}^{(g)} \vc{g}} = 1.\\
\end{aligned}
\end{equation}
We show examples of the zero-padded cost matrices, using the two example motion priors, in Figure~\ref{fig:cost-matrices}.

\paragraph{Certificate}

Comparing (Q-GP) with (Q) it is clear that a new certificate can be derived by checking conditions~\eqref{eq:primal-feas} to~\eqref{eq:stationarity}, but with the new system matrices. In particular, we define the new certificate matrix
\begin{equation}
  \vc{H}_{GP}(\hat{\rho}, \hc{\lambda}) = \fracE \vc{Q}^{(g)} + \fracN \vc{R}^{(g)} + \hat{\rho}\vc{A}_0^{(g)} + \sum_n \hat{\lambda}_n \vc{A}_n^{(g)}. 
\end{equation}
We note again that primal feasibility~\eqref{eq:primal-feas} is given by construction. The stationarity condition~\eqref{eq:stationarity} now reads
\begin{equation}\label{eq:lin-sys}
  \bvec{\vc{A}_1^{(g)}\hc{g} & \cdots & \vc{A}_N^{(g)}\hc{g} & \vc{A}_0^{(g)}\hc{g}} \hc{y} = -\fracE\vc{Q}^{(g)}\hc{g} -\fracN\vc{R}^{(g)}\hc{g}.
\end{equation}
\cor{As in Section~\ref{sec:methods-multi}, we show in \corn{the supplementary material}~\cite{safe_arxiv} 
that this system of equations has a unique solution, given by~\eqref{eq:lambda} for $\hc{\lambda}$ and 
\begin{align}
  \hat{\rho} &= - \fracE\sum_n \vc{e}_n^\top\Cov_n^{-1}\vc{e}_n - \fracN \hc{\theta}^\top \vc{R} \hc{\theta}.  \label{eq:rho-gp-here}
\end{align}
\noindent
Again,} if the dual variables are feasible, then~\eqref{eq:rho-gp-here} implies strong duality.

\subsection{Iterative Solver}\label{sec:solver}

The proposed certificate is applicable to any solution candidate satisfying first-order stationarity. A number of nonlinear least-squares solvers could be used to obtain such a candidate~\cite{nocedal_numerical_2006}. We give a brief outline of our implemented sparse \ac{GN} solver, which we choose because of its wide usage and efficiency.
\cor{Linearizing the least-squares residuals around a current estimate $\vc\theta^{k}$, the optimal update $\delta\vc{\theta}$ is the solution of}
\begin{equation}
  \begin{aligned}
    \left(\vc{R} + \cor{\frac{N}{E}} \vc{J}^\top\vc{\Sigma}^{-1}\vc{J}\right)&\delta\vc{\theta} = 
  - \vc{R}\vc{\theta}^k
  + \cor{\frac{N}{E}} \vc{J}^\top\vc{\Sigma}^{-1}\left(\cor{\vc{d}} - \vc{h}(\vc{\theta}^k)\right).\label{eq:GN}
  \end{aligned}
\end{equation}
Here, $\vc{J} := \Diag{\left(\vc{\nabla}_{\vc{x}_n}\vc{h}_n\right)}_{n=1}^N$ is the measurement Jacobian, and $\vc{\Sigma}^{-1} :=\Diag{\left(\Cov_n^{-1}\right)}_{n=1}^N$. The vector $\vc{d}$ contains all stacked distance measurements $\vc{d}^\top=[\vc{d}_1^\top \cdots \vc{d}_N^\top]$. \corn{We refer the reader to~\cite[4.4.3]{barfoot_state-estimation} for a detailed treatment of \ac{GN} for continuous-time estimation.}
\cornn{We stop the algorithm when the root-mean-squared step size is less than $10^{-10}$.} Equation~\eqref{eq:GN} is a sparse linear system of equations due to the form of the left-hand-side matrices, and can be solved efficiently via, for instance, sparse Cholesky factorization. The complexity of each iteration is thus $O(N)$ and we found that in practice, convergence \corn{usually takes less than 10} iterations.

\subsection{Efficient Certificate Computation}\label{sec:sparsity}

For a practical solution, we require not only an efficient iterative solver, but also an efficient certificate. Since we can solve analytically for the optimal dual variables in $O(N)$ time, the bottleneck of the computation lies in certifying \ac{psd}-ness of the certificate matrix, which is of size $N(K+1) + 1$. The most intuitive approach of computing the eigenvalues of this matrix is prohibitively expensive, with complexity of up to $O(N^3)$~\cite{golub_matrix_2013}. Thankfully, we can exploit the particular sparsity pattern of the matrix to bring the cost of the certificate down to $O(N)$, as we will outline next. 

The certificate matrix is a block-tridiagonal arrowhead matrix and belongs to the class of chordally sparse matrices, which exhibit numerous interesting properties (see~\cite{vandenberghe_chordal_2015} for an overview). The chordal property we exploit here is that the sparsity pattern is `preserved' in the $\vc{L}$ matrix of the $\vc{LDL}^\top$ decomposition. For our sparsity pattern, the certificate matrix is \ac{psd}, if and only if it can be decomposed as
$\vc{H} = \vc{L}\vc{D}\vc{L}^\top$, with $\vc{D} = \Diag([(\vc{D}_n)_{n=1}^{N}, \delta])$, and 
\begin{equation}
  \begin{aligned}
    \vc{L} &= \bvec{
      \vc{J}_{1} & \vc{0}   & \smash{\cdots} & \smash{\cdots} & \vc{0} \\
      \vc{L}_{1} &  \smash{\ddots} & \smash{\ddots} & & \smash{\vdots} \\
      \smash{\vdots} & \smash{\ddots} & \smash{\ddots} & \smash{\ddots} & \smash{\vdots} \\
      \vc{0} & \smash{\cdots} & \vc{L}_{N\- 1} & \vc{J}_N & \vc{0}\\
    \vc{l}_1^\top &  \smash{\cdots} & \vc{l}_{N\- 1}^\top & \vc{l}_N ^\top & 1 
    },
\end{aligned}
\label{eq:l}
\end{equation}
where $\vc{J}_n$ are lower-diagonal matrices and $\vc{D}_n$ are diagonal matrices with non-negative elements. $\vc{L}_n$ and $\vc{l}_n$ are a priori dense matrices and vectors, respectively. Equating the non-zero elements in~\eqref{eq:l} with the corresponding blocks of $\vc{H}(\hat{\rho},\hc{\lambda})$, we obtain the following equalities:
  \begin{align}
    \vc{H}_{nn} &= \begin{cases}
      \ul{\vc{J}_n} \ul{\vc{D}_n} \ul{\vc{J}_n} ^\top &\text{for }n=1 \\
      \vc{L}_{n\- 1}\vc{D}_{n\- 1}\vc{L}_{n\- 1}^\top + 
      \ul{\vc{J}_n} \ul{\vc{D}_n} \ul{\vc{J}_n} ^\top &\text{for }2\leq n \leq N \\
    \end{cases}, \nonumber \\
    \vc{h}_n &= \begin{cases}
      \vc{J}_n\vc{D}_n\ul{\vc{l}_n} &\text{for }n=1 \\ 
      \vc{L}_{n\- 1}\vc{D}_{n\- 1}\vc{l}_{n\- 1} + \vc{J}_n\vc{D}_n\ul{\vc{l}_n}
      &\text{for }2\leq n \leq N
    \end{cases},\nonumber \\ 
    \vc{H}_{n, n+1} &= \vc{J}_{n}\vc{D}_{n}\ul{\vc{L}_{n}}^\top \quad \text{for } 1\leq n \leq N-1, \nonumber \\
    h &= \sum_{n=1}^N \vc{l}_n^\top \vc{D}_n \vc{l}_n + \ul{\delta}. \label{eq:ldl-algo}
\end{align}
These equations define a recursive scheme for computing the decomposition; the unknown factors can be computed in the order $\vc{D}_1, \vc{J}_1, \vc{L}_1, \ldots, \vc{D}_N, \vc{J}_N, \vc{L}_N, \vc{l}_1,\ldots,\vc{l}_N, \delta$. The factors $\vc{D}_n$ and $\vc{J}_n$ are computed through individual $\vc{LDL}^\top$ decompositions, but the involved matrices are only of size $K+1$. We can stop computing the decomposition early if we find a negative diagonal value, as the certificate has failed. The algorithm runs in $O(N)$ and as we show in simulation, has a similar absolute runtime as the \ac{GN} solver\footnote{\cor{In practice, the decomposition may suffer from numerical instability~\cite[Section 4.2]{golub_matrix_2013}, meaning that when the smallest eigenvalue of $\vc{H}$ is slightly negative, the diagonal elements of $\bm{D}$ may become strongly negative, wrongly indicating a failed certificate. To mitigate this fact, we add a small regularization $\bm{H}\leftarrow \bm{H}+\beta \bm{I}$ before computing the decomposition. We set the value to $\beta=1e^{-3}$ throughout the real-world experiments. However, preliminary experiments suggest that different anchor configurations may require a different threshold. Automatic tuning of $\beta$ or a more numerically stable certification method are subjects of current investigation.}}.

\section{Experimental Results}\label{sec:results}

In this section, we show the effectiveness of the proposed method \cor{in simulation and on a real-world dataset}. We first study the certificate in simulation, showing that our local solver finds the optimal solution in the majority of cases for random setups, and when it fails to do so, the certificate does not hold \corn{for all but the highest considered noise levels. Incorrect local solutions are typically only found for few random initializations, suggesting that in practice, randomly reinitializing until the certificate holds is a viable strategy.}
Next, we use the certificate to evaluate the localization performance in a real-world scenario with distance measurements from \ac{UWB} anchors measured on a drone and show that we can successfully differentiate local and global solutions\footnote{\cornn{The Python code to reproduce all results is available at \url{https://github.com/utiasASRL/safe_and_smooth}.}}.

\begin{figure}[t]
    \centering
    \includegraphics[width=\linewidth]{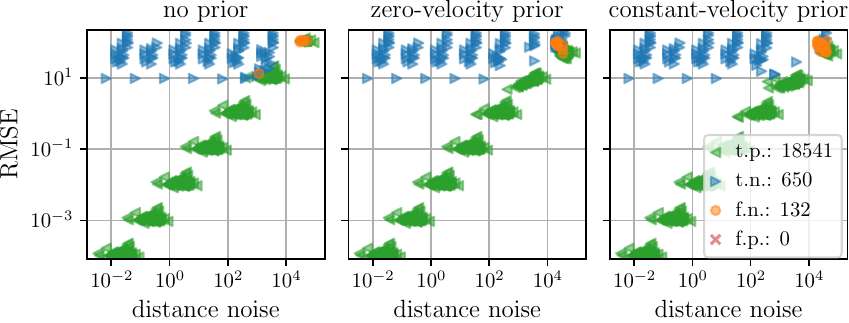}
    \caption{Certificate value vs. \ac{rmse}, using different motion priors, \cor{in} simulations with $N=100$, $M=6$ and $D=2$. The group of solutions corresponding to the smallest \cor{cost} out of \cor{10} initializations are labelled optimal. \cor{Disregarding the highest noise levels, all optimal solutions are successfully identified (true positives, t.p.) and the certificate fails for suboptimal solutions (true negatives, t.n.). At the highest noise levels, a small proportion of optimal solutions are not certified (false negatives, f.n.), which is in line with the sufficiency of the certificate.}}
    \label{fig:certificate-simulated}
\end{figure}

\subsection{Simulation Results}\label{sec:simulation}

\begin{figure}[t]
    \centering
    \includegraphics[width=\linewidth]{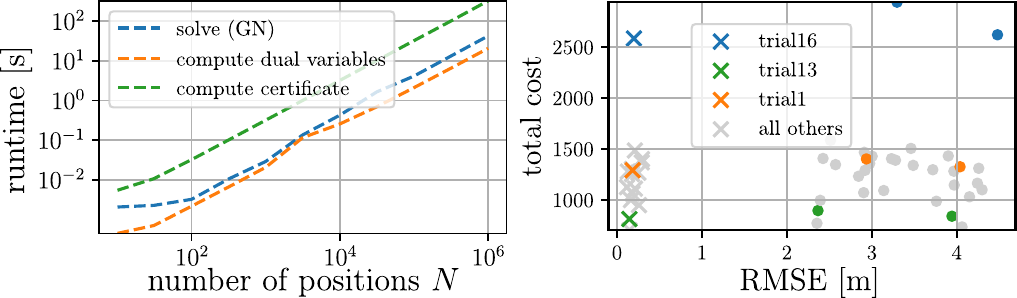}
    \caption{Left: Computation time of our \ac{GN} solver, evaluating the dual variables, and computing the certificate, respectively, with increasing number of positions $N$. \cor{Right: certificate evaluation on real data, comparing the final cost and \ac{rmse} of solutions. Three solutions for three chosen datasets are highlighted in colors. For all datasets, the cost difference between certified solutions (cross) and uncertified solutions (circle) is small, but the resulting \ac{rmse} difference is significant. Thanks to the proposed method, such suboptimal solutions can be avoided.}}
    \label{fig:certificate-time-rmse}
\end{figure}

\mytitle{Setup} We create \corn{2D simulated experiments by generating} trajectories according to the constant-velocity model, where for each random instance we draw an initial velocity and position vector uniformly and \corn{coordinate-wise} from $[-1, 1]$. \lt{Note that we consider all dimensions normalized thus unit-less.} The noise on the acceleration is \corn{assumed independent with $\sigma_a=0.2$, so that $\vc{Q}_C=\sigma_a\vc{I}$}. The \cor{anchor coordinates} are also drawn uniformly at random from $[0, 1]$ and their bounding box is scaled to the size of the trajectory. We set $N=100$ and $M=6$. We measure distances from all anchors at each time, and we generate measurements by adding Gaussian noise to the true distances before squaring them. We assume \emph{i.i.d.} zero-mean noise with variance $\sigma_d$. For each random experimental setup, we solve using the \ac{GN} algorithm, using \corn{10} different random initializations, and using no motion prior, a zero-velocity motion prior and a constant-velocity motion prior, respectively. We set $\vc{\Sigma}_n$ and $\vc{Q}_C$ to the true values.

\mytitle{Results} \corn{First, we} study the effect of noise in a quantitative analysis in Figure~\ref{fig:certificate-simulated}. We vary the \cor{measurement} noise $\sigma_d$ and report the \ac{rmse} between the estimated and ground-truth trajectory as a function of the \corn{effective distance noise}. Since we lack an optimal solver, we label solutions as globally optimal when they correspond to the smallest \cor{cost} (up to numerical tolerance) for a given setup, and locally optimal otherwise. This method is reliable for small noise levels, as a big gap exists between the \cor{cost} of the local and global solutions, but less so for higher noise. \cor{Using this method, we can identify that the majority of solutions (96\%) are true positives --- global solutions \lt{where} the certificate holds. Even more importantly, we observe that there are no false positives --- the certificate never holds for a suboptimal solution. Out of the uncertified solutions, 20\% are false negatives, meaning the certificate misses an optimal solution. However, this happens only at the highest noise levels considered,} \corn{suggesting the existence of a high noise threshold up to which strong duality holds. For lower noise levels, we conclude that} although the certificate is only a sufficient condition in theory, it is effectively a necessary condition in practice: when the certificate does not hold, the solution \cor{is usually} suboptimal. \cor{We also note that the method and certificate are robust to model mismatch: although the real trajectory is of the constant-velocity type, all motion priors yield satisfactory results. \cornn{A study of the setups leading to local optima, given in the supplementary material~\cite{safe_arxiv}, reveals that local minima are usually the result of poor (\ie, almost co-linear) anchor placement}.} 

We also study the computation time with increasing number of positions $N$. The left plot of Figure~\ref{fig:certificate-time-rmse} shows that both the computation times of the solver and certificate increase linearly \new{with $N$}. \cornn{While the certificate takes slightly longer than \ac{GN} on average, this difference may be reduced by a more efficient implementation}. We run our method on up to a million states \corn{(using the highest-dimensional constant-velocity prior)}, \corn{underlining} the scalability of the approach. 

\subsection{Experimental UWB Drone Dataset}\label{sec:experiments}

\begin{figure}[t]
    \centering
    \includegraphics[width=\linewidth]{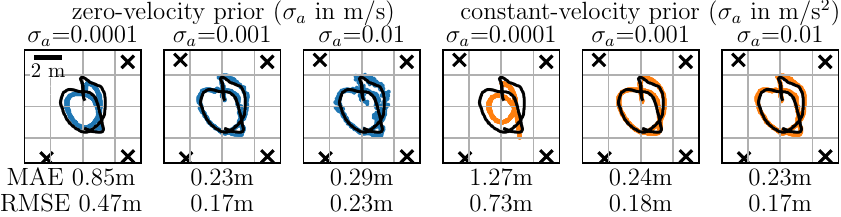}
    \caption{Planar projections of certified solutions for the first flying drone dataset, using the zero-velocity prior (blue) and the constant-velocity prior (orange), with varying parameter $\sigma_a$. The ground truth is shown in black, and the \ac{rmse} and \ac{mae} are shown below each plot. \corn{The motion prior has to strike a balance between over-smoothing and over-fitting to noisy measurements. We find that the constant-velocity prior with $\sigma_a=1e^{-3}$ m$/$s$^2$ yields consistent results for all datasets.}}
    \label{fig:real-solutions}
\end{figure}

Finally, we test the certificate on a dataset collected by a drone in a flying arena. \corn{We use $M=4$ \ac{UWB} anchors, placed in the 4 top corners} of an arena, which is about 7 by 7 meters wide and 3 meters high. Measurements are obtained asynchronously, resulting in exactly one distance measurement at a time. Ground truth position is obtained with a \textit{Vicon} motion capture setup. The setup is described in more detail in~\cite{goudar_gaussian_2022}. In total, 16 different flights (trials), are performed, with different trajectory characteristics and varying velocities, \lt{using a constant-acceleration policy}.
The average measurement frequency per anchor is \SI{5}{Hz}, and each trial being a bit more than one minute long, we obtain about $N=1600$ measurement times for each trial. \corn{The average runtime of our estimation algorithm including certificate is \SI{5}{s}\lt{, of which \SI{3.5}{s} are spent on certification.}} \corn{The bias in the distance measurements, typical for \ac{UWB} measurements, is removed in a prior calibration phase.}

Since the motion model is unknown a priori, we perform calibration on the first dataset to identify suitable values for $\sigma_a$ (we fix $\sigma_d$ to \SI{5}{cm} according to the expected accuracy of the \ac{UWB} anchors). Figure~\ref{fig:real-solutions} shows the estimates obtained using different combinations of $\sigma_a$ and motion priors;  \lt{We use the constant-velocity prior and $\sigma_a=1e^{-3}$ m$/$s$^2$}, for a good balance between smoothing and overfitting to noise, which is reflected in a low \ac{rmse} and \ac{mae}, respectively.

\corn{Figure~\ref{fig:local-optimum-real} shows two example estimates for the first dataset, both of which correspond to stationary points of the problem, with one corresponding to the global minimum and one to a local minimum with high \ac{rmse}. The certificate successfully identifies the global minimum, and it does not hold for the local minimum. The same behavior is observed on all datasets, which are summarized in Figure~\ref{fig:certificate-time-rmse}. Note that all certified solutions, marked with crosses, correspond to a significantly lower \ac{rmse} than the local solutions, for which the certificate does not hold. The cost, on the other hand, is quite similar between global and local minima (three examples are highlighted in Figure~\ref{fig:certificate-time-rmse}), which suggests that it cannot be used for automatically evaluating optimality.} \cornn{Like in the simulated measurements, we observe that the local solution seems to be the result of poor anchor placement: the anchors are almost co-planar. Indeed, when using measurements from two additional ground-based anchors, it was found that the local solution always converges to the global optimum.}

\section{Conclusion and Future Work}\label{sec:conclusion}

We have provided optimality certificates for range-only continuous-time localization. 
We have derived a closed-form solution for the optimal Lagrange multipliers that depends only on the residuals of the problem, and provided an efficient method for checking if the certificate matrix is~\ac{psd}. We have successfully certified the solutions found by a sparsity-exploiting \ac{GN} solver both in simulation and real experiments, and observed that the global optimum is found in most cases, in particular when the anchors are placed in non-degenerate configurations. We hope that the proposed certificate is a first step to extend existing provably optimal solvers to distance and other nonlinear measurements. A promising line of future work is the extension of the results to the full~\ac{SLAM} setup, where the anchor positions are unknown a priori. In a second step, the sparsity of the problem suggests that incorporating certificates into incremental solvers such as~\cite{Kaess2012} could decrease the computational cost of the proposed method even further. Finally, the least-squares formulation used herein is sensitive to outliers, and could be replaced with a more robust cost function~\cite{yang_teaser_2020,fineline}. 

\vspace{-0.1em}
\section*{Acknowledgments}

We would like to thank Abhishek Goudar of the Dynamic Systems Lab for providing us with the drone dataset. 

\vspace{-0.2em}
\bibliographystyle{IEEEtran}
\bibliography{zotero-final}

\appendix
This appendix is published along with the \textit{arXiv} verison of this paper and provides additional mathematical details and results.

\subsection{Derivation of Dual Variables}

\newcommand{\Sig}{\ensuremath{\bm{\Sigma}^{-1}}}

We provide the detailed derivations for the computation of the optimal dual variables $\hat{\rho}$ and $\hc{\lambda}$, given a candidate solution $\hc{f}$. First, plugging \corr{in} the expressions for $\vc{Q}$, $\vc{A}_i$ ($i=0\ldots N$), $\vc{R}$ and $\hc{f}$ in~\eqref{eq:lin-sys-here}, \corr{we obtain}
\corr{
\begin{equation}
\begin{aligned}
  \begin{bmatrix}
    \hat{\lambda}_1\hc{x}_1 \\
    -\frac{\hat{\lambda}_1}{2} \\
    \hat{\lambda}_2\hc{x}_2 \\
    -\frac{\hat{\lambda}_2}{2} \\
    \vdots \\
    \hat{\rho} - \sum_n\frac{\hat{\lambda}_n}{2}\norm{\hc{x}_n}^2
  \end{bmatrix} 
  &= \frac{1}{E}\begin{bmatrix}
  -2\vc{Y}_1\Sig_1 \vc{e}_1 \\
  \vc{1}^\top\Sig_1 \vc{e}_1 + \\
  -2\vc{Y}_2\Sig_2 \vc{e}_2 \\
  \vc{1}^\top\Sig_2 \vc{e}_2\\
  \vdots \\ 
  -\sum_n \vc{b}_n^\top\Sig_n \vc{e}_n \\
\end{bmatrix}.
\end{aligned}
\end{equation}
\noindent
The} $N$ equations corresponding to the substitution variables take the form
\begin{equation}
  \foralln -\frac{1}{2}\hat{\lambda}_n =  \corr{\frac{1}{E}}\vc{1}^\top \Cov_n^{-1}\vc{e}_n ,
  \label{eqapp:lambda}
\end{equation}
which can be solved for $\hat{\lambda}_n$. Plugging the solution into the other $ND$ rows involving $\hat{\lambda}_n$, we need to show that
\begin{equation}
  \foralln \hc{x}_n\hat{\lambda}_n = -2 \corr{\frac{1}{E}} \hc{x}_n\vc{1}^\top \Cov_n^{-1}\vc{e}_n \stackrel{?}{=} -2
  \corr{\frac{1}{E}} \anchors_n\Cov_n^{-1}\vc{e}_n,
  \label{eq:lin-sys2}
\end{equation}
or in other words, we need these equations to be redundant. Because $\hc{x}_n$ are stationary points of~\eqref{eq:cost-ro}, we have  
\begin{equation}\label{eq:grad-f}
  \foralln \vc{\nabla}_{\vc{x}_n}f =  
  4(\anchors_n - \hc{x}_n\vc{1}^\top)\Cov_n^{-1}\vc{e}_n  = \vc{0},
\end{equation}
from which~\eqref{eq:lin-sys2} follows trivially.  Finally, we can use the last row of the linear system in~\eqref{eq:lin-sys-here} to solve for $\hat{\rho}$, which gives
\begin{equation}
\begin{aligned}
  \hat{\rho} &= \sum_n \left(\frac{\hat{\lambda}_n}{2}\norm{\hc{x}_n}^2 
  - \corr{\frac{1}{E}} \vc{b}_n^\top\Cov_n^{-1}\vc{e}_n\right) \\
  &= -\corr{\frac{1}{E}} \sum_n \left(\norm{\hc{x}_n}^2\vc{1}^\top + \vc{b}_n^\top\right)\Cov_n^{-1}\vc{e}_n   \\
  &= -\corr{\frac{1}{E}}\sum_n \left(\vc{e}_n^\top - 2\hc{x}_n^\top\anchors_n + 2\norm{\hc{x}_n}^2\vc{1}^\top\right)\Cov_n^{-1}\vc{e}_n  \\
  &= -\corr{\frac{1}{E}}\sum_n \vc{e}_n^\top\Cov_n^{-1}\vc{e}_n - 2\hc{x}_n^\top\left(\anchors_n - \hc{x}_n\vc{1}^\top\right)\Cov_n^{-1}\vc{e}_n  \\
  &= -\corr{\frac{1}{E}}\sum_n \vc{e}_n^\top\Cov_n^{-1}\vc{e}_n.
\end{aligned}\label{eqapp:rho}
\end{equation}
where we have substituted in $\hat{\lambda}_n$ in the first step, the definition of $\vc{b}_n$ in the second step, and the stationarity condition~\eqref{eq:grad-f} to yield the final expression.

\subsubsection{GP solution}~\label{app:gp} \corr{Building on the result from the previous section, we can derive the form of the optimal dual variables when we add a regularization term to the cost function. Starting from~\eqref{eq:lin-sys}, we arrive at almost the same system of equations as before, but with additional rows for the new dimensions in $\vc{\theta}$, and additional terms for the motion prior, added on the right-hand side:}
\corr{\begin{equation*}\label{eq:lin-sys-3}
\begin{aligned}
  \begin{bmatrix}
    \hat{\lambda}_1\hc{x}_1 \\
    \vc{0} \\
    -\frac{\hat{\lambda}_1}{2} \\
    \hat{\lambda}_2\hc{x}_2 \\
    \vc{0} \\
    -\frac{\hat{\lambda}_2}{2} \\
    \vdots \\
    \hat{\rho} - \sum_n \frac{\hat{\lambda}_n}{2}\norm{\hc{x}_n}^2
  \end{bmatrix} 
&= 
\frac{1}{E}
\begin{bmatrix}
    -2\vc{Y}_1\Sig_1 \vc{e}_1  \\
    \vc{0} \\
  \vc{1}^\top\Sig_1 \vc{e}_1 \\
   -2\vc{Y}_2\Sig_2 \vc{e}_2 \\
   \vc{0} \\
  \vc{1}^\top\Sig_2 \vc{e}_2 \\
  \vdots \\ 
  -\sum_n \vc{b}_n^\top\Sig_n \vc{e}_n \\
\end{bmatrix}
+ \frac{1}{N}
\begin{bmatrix}
 \vc{R}_{1,x}\hc{f} \\
\vc{R}_{1,v}\hc{f} \\ 
0 \\
 \vc{R}_{2,x}\hc{f} \\
\vc{R}_{2,v}\hc{f} \\ 
0 \\
\vdots \\ 
0
 \end{bmatrix}. 
\end{aligned}
\end{equation*}
$\vc{R}_{n,x}$ and }$\vc{R}_{n,v}$ are the first $D$ rows, and remaining rows, respectively, of $\vc{R}$ as defined in~\eqref{eq:R}. 

Note that \corn{both the left-hand and right-hand sides do not change for the rows corresponding to the substitutions}. Therefore, we can still use~\eqref{eqapp:lambda} to solve for $\hc{\lambda}$. \corr{The last step is to show that the remaining rows are redundant, which we show for the two example motion priors separately.} 

\corr{
  \textbf{Example 1:} For the zero-velocity prior, we have $\hc{\theta}=\hc{x}$ and the rows with $\vc{R}_{n,v}$ in~\eqref{eq:lin-sys-3} vanish. Then, we only need to show that
\begin{equation}
  \foralln \hc{\theta}_n\hat{\lambda}_n 
  = \hc{x}_n \corr{\hat{\lambda}_n}
  \stackrel{?}{=}2\anchors_n\Cov_n^{-1}\vc{e}_n + \frac{1}{N} \vc{R}_{n,x} \hc{x}.
  \label{eq:lambda-test-zero}
\end{equation}
Equations~\eqref{eq:lambda-test-zero} hold because $\hc{\theta}_n$ are stationary points of the cost function $f(\vc{\theta}) + r(\vc{\theta})$ defined in~\eqref{eq:cost-gp}, which means that
\begin{align}
  \vc{0} &= \vc{\nabla}_{\vc{x}_n}f(\hc{\theta}) 
  + \vc{\nabla}_{\vc{x}_n}r(\hc{\theta}) \\
&= 4\corr{\frac{1}{E}}(\anchors_n - \hc{x}_n\vc{1}^\top)\Cov_n^{-1}\vc{e}_n
+ 2\corr{\frac{1}{N}}\vc{R}_{n, x} \hc{x}, \label{eq:grad_x}
\end{align}
and~\eqref{eq:lambda-test-zero} follows.
}

\textbf{Example 2:} \corr{For the constant-velocity motion prior, we need to show that }
\begin{equation}
  \foralln \hc{\theta}_n\hat{\lambda}_n 
  = \bvec{\hc{x}_n \\ \hc{v}_n} \corr{\hat{\lambda}_n}
  \stackrel{?}{=}\bvec{2\anchors_n\Cov_n^{-1}\vc{e}_n \\ \vc{0}} + \bvec{\vc{R}_{n,x} \\ \vc{R}_{n,v}} \hc{\theta}.
  \label{eq:lambda-test}
\end{equation}
As for the first example, the first $D$ rows of system~\eqref{eq:lambda-test} hold because $\hc{\theta}_n$ are stationary points with respect to $\vc{x}_n$, so~\eqref{eq:grad_x} must hold. For the last $D$ rows, we have
\begin{align}
  \vc{0} &= \vc{\nabla}_{\vc{v}_n}r(\hc{\theta}) 
  = \corr{\frac{1}{N}}\vc{R}_{n, v} \hc{\theta} 
  \label{eq:gradient-v}, 
\end{align}
which confirms that the additional \lt{equations} are satisfied.
 
\noindent
Finally, we start with the same expression for $\hat{\rho}$ as in \corr{the second-last row of~\eqref{eqapp:rho}}, but this time, using~\eqref{eq:gradient-v}, we obtain
\begin{equation}
\begin{aligned}
  \hat{\rho} 
  &= -\corr{\frac{1}{E}}\sum_n \vc{e}_n^\top\Cov_n^{-1}\vc{e}_n - 2 \hc{x}_n^\top\left(\anchors_n - \hc{x}_n\vc{1}^\top\right)\Cov_n^{-1}\vc{e}_n  \\
  &= - \corr{\frac{1}{E}} \sum_n \vc{e}_n^\top\Cov_n^{-1}\vc{e}_n -\corr{\frac{1}{N}} \sum_n \hc{x}_n^\top \vc{R}_{n,x} \hc{\theta} \\
  &= - \corr{\frac{1}{E}} \sum_n \vc{e}_n^\top\Cov_n^{-1}\vc{e}_n -\corr{\frac{1}{N}} \sum_n \hc{x}_n^\top \vc{R}_{n,x} \hc{\theta} - \hc{v}_n^\top \vc{R}_{n,v} \hc{\theta} \\
  &= - \corr{\frac{1}{E}} \sum_n \vc{e}_n^\top\Cov_n^{-1}\vc{e}_n -\corr{\frac{1}{N}} \hc{\theta}^\top \vc{R} \hc{\theta},
\end{aligned}\label{eq:rho-gp}
\end{equation}
where we have used~\eqref{eq:grad_x} for the second line. From there, we add  $\sum_n \hc{v}_{n}^\top \vc{R}_{n,v} \hc{\theta}$, which is zero because of~\eqref{eq:gradient-v}, to make the relationship with the quadratic cost function more evident.

\subsection{Simulation results}

\begin{figure}[tb]
  \centering
  \includegraphics[width=\linewidth]{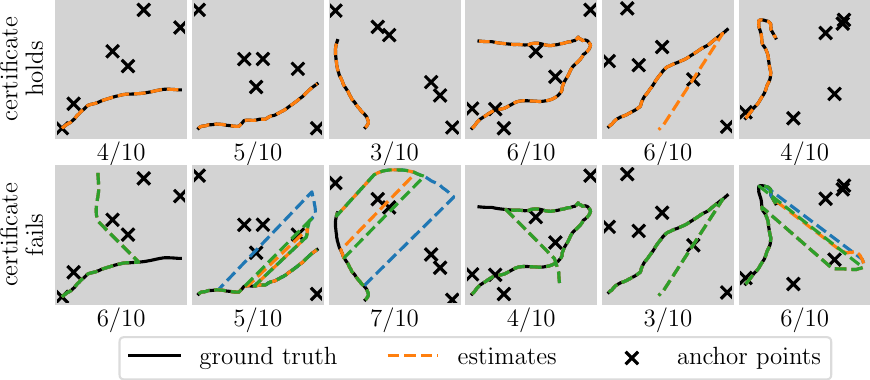}
  \caption{Visualization of \cor{optimal (top) and suboptimal (bottom)} solutions found by a local \ac{GN} solver in range-only continuous-time localization\lt{, in simulation}. \cor{Each column corresponds to a different random setup.} \new{Solid \cor{black} lines correspond to ground truth trajectories and the dashed coloured lines \cor{correspond} to estimates from various random initializations.} Out of 100 random \cor{setups}, the 6 shown \cor{setups} yield, for a high proportion of \cor{random} initializations (see labels below each plot), suboptimal solutions. The proposed method allows to identify such suboptimal solutions with little additional computational cost.} 
  \label{fig:local-optimum}
\end{figure}

\corn{In Figure~\ref{fig:local-optimum}, we study the simulated setups \corn{prone to} yield local optima. We fix the noise level to $\sigma_d=10^{-3}$ and show the results for the constant-velocity prior. A qualitative analysis suggests that, whenever the anchors are sufficiently spread, meaning they are not almost co-linear (or almost co-planar in the three-dimensional case), the local solver converges to the global minimum. Indeed, for the shown $6$ out of $100$ random setups, which are the only ones consistently leading to local minima, the anchors are close to co-linear. The local optima, shown in the second row of Figure~\ref{fig:local-optimum}, are partially mirrored versions of the optimal solution, around the line defined by the anchors. For each setup, we show three local solutions in dashed lines of different colors. \lt{The proportion of local and global solutions, respectively, is shown below each plot}. At this noise level, the certificate correctly labels all optimal solutions, and fails for all local solutions. As noted in Section~\ref{sec:simulation}, this suggests that strong duality holds and that the certificate is sufficient \textit{and} necessary, for this noise level.}

\end{document}